\documentclass[10pt,twocolumn,letterpaper]{article}

\usepackage[pagenumbers]{cvpr} 

\usepackage{graphicx}
\usepackage{amsmath}
\usepackage{amssymb}
\usepackage{booktabs}
\usepackage[table]{xcolor}
\usepackage{multicol}
\usepackage{multirow}
\usepackage{footnote}
\usepackage{makecell}
\usepackage{CJKutf8}
\definecolor{darkgreen}{RGB}{35,140,0}
\definecolor{darkred}{RGB}{160,35,0}
\newcommand\blfootnote[1]{%
\begingroup 
\renewcommand\thefootnote{}\footnote{#1}%
\addtocounter{footnote}{-1}%
\endgroup 
}

\definecolor{Klein_Blue}{rgb}{0.0, 0.129, 0.6}
\usepackage[pagebackref,breaklinks,colorlinks]{hyperref}
\hypersetup{
    colorlinks=true,
    linkcolor=magenta,
    urlcolor=magenta,
    citecolor=Klein_Blue
    }

\usepackage[capitalize]{cleveref}
\crefname{section}{Sec.}{Secs.}
\Crefname{section}{Section}{Sections}
\Crefname{table}{Table}{Tables}
\crefname{table}{Tab.}{Tabs.}

\def\cvprPaperID{} 
\def\confName{CVPR}
\def\confYear{2023}

\begin{document}

\title{EfficientViT: Memory Efficient Vision Transformer with \\ Cascaded Group Attention}
    \author{Xinyu Liu$^{1, *}$ , Houwen Peng$^{2}$, Ningxin Zheng$^{2}$, Yuqing Yang$^{2}$, Han Hu$^{2}$, Yixuan Yuan$^{1}$\\ 
    $^1$ The Chinese University of Hong Kong,  $^2$ Microsoft Research\\
    \newline
    }
\maketitle

\begin{abstract}
\vspace{-8pt}
    Vision transformers have shown great success due to their high model capabilities. However, their remarkable performance is accompanied by heavy computation costs, which makes them unsuitable for real-time applications. 
    In this paper, we propose a family of high-speed vision transformers named EfficientViT. 
    We find that the speed of existing transformer models is commonly bounded by memory inefficient operations, especially the tensor reshaping and element-wise functions in MHSA. Therefore, we design a new building block with a sandwich layout, i.e., using a single memory-bound MHSA between efficient FFN layers, which improves memory efficiency while enhancing channel communication. Moreover, we discover that the attention maps share high similarities across heads, leading to computational redundancy. To address this, we present a cascaded group attention module feeding attention heads with different splits of the full feature, which not only saves computation cost but also improves  attention diversity. Comprehensive experiments demonstrate EfficientViT outperforms existing efficient models, striking a good trade-off between speed and accuracy. For instance, our EfficientViT-M5 surpasses MobileNetV3-Large by 1.9\% in accuracy, while getting 40.4\% and 45.2\% higher throughput on Nvidia V100 GPU and Intel Xeon CPU, respectively. Compared to the recent efficient model MobileViT-XXS, EfficientViT-M2 achieves 1.8\% superior accuracy, while running 5.8×/3.7× faster on the GPU/CPU, and 7.4× faster when converted to ONNX format. Code and models are available at \href{https://github.com/microsoft/Cream/tree/main/EfficientViT}{here}.
\end{abstract}

\vspace{-8pt}
\blfootnote{
  $^*$Work done when Xinyu was an intern of Microsoft Research.
 }
\vspace{-19pt}
\section{Introduction}
\label{sec:intro}

\vspace{-4pt}
\begin{figure}
\vspace{-8pt}
    \centering
    \includegraphics[width=0.38\textwidth]{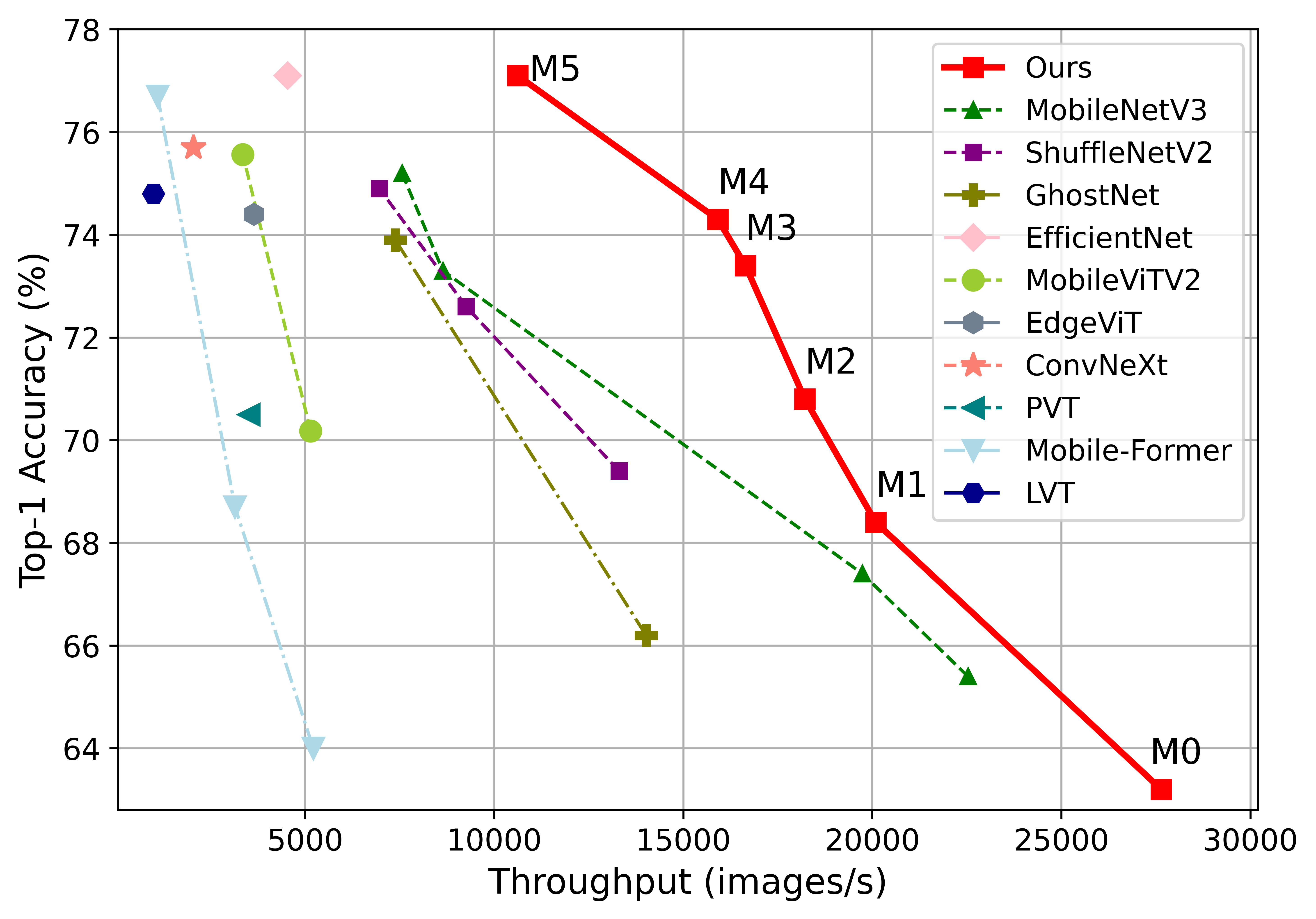}
    \vspace{-10pt}
    \caption{Speed and accuracy comparisons between EfficientViT (Ours) and other efficient CNN and ViT models tested on an Nvidia V100 GPU with ImageNet-1K dataset \cite{imagenet}.}
    \label{fig:model_gpu}
    \vspace{-18pt}
\end{figure}

Vision Transformers (ViTs) have taken computer vision domain by storm due to their high model capabilities and superior performance \cite{ViT, deit, swin}.
However, the constantly improved accuracy comes at the cost of increasing model sizes and computation overhead. For example, SwinV2 \cite{swin_v2} uses 3.0B parameters, while V-MoE \cite{V-MoE} taking 14.7B parameters, to achieve state-of-the-art performance on ImageNet~\cite{imagenet}. 
Such large model sizes and the accompanying heavy computational costs make these models unsuitable for applications with real-time requirements\cite{MiniViT, li2022efficientformer, wu2022tinyvit}.

There are several recent works designing light and efficient vision transformer models 
\cite{lvt, mobileformer, mobilevit, nasvit, pan2022edgevits,  maaz2022edgenext, pale, huang2022lightvit}.
Unfortunately, most of these methods aim to reduce model parameters or Flops, which are indirect metrics for speed and do not reflect the actual inference throughput of models. 
For example, {MobileViT-XS} \cite{mobilevit} using 700M Flops runs much slower than DeiT-T \cite{deit} with 1,220M Flops on an Nvidia V100 GPU.
Although these methods have achieved good performance with fewer Flops or parameters, many of them do not show significant wall-clock speedup against standard isomorphic or hierarchical transformers, \emph{e.g.}, DeiT \cite{deit} and Swin \cite{swin},
and have not gained wide adoption. 

To address this issue, in this paper, we explore how to go faster with vision transformers, seeking to find principles for designing efficient transformer architectures. Based on the prevailing vision transformers DeiT \cite{deit} and Swin \cite{swin}, we systematically analyze three main factors that affect model inference speed, including memory access, computation redundancy, and parameter usage. In particular, we find that the speed of transformer models is commonly {memory-bound}. In other words, memory accessing delay prohibits the full utilization of the computing power in GPU/CPUs \cite{gu2021towardsmemory, jiang2020characterizing, venkat2019swirl}, leading to a critically negative impact on the runtime speed of transformers \cite{ivanov2021datamovement, dao2022flashattention}. The most memory-inefficient operations are the frequent tensor reshaping and element-wise functions in multi-head self-attention (MHSA). We observe that through an appropriate adjustment of the ratio between MHSA and FFN (feed-forward network) layers, the memory access time can be reduced significantly without compromising the performance. 
Moreover, we find that some attention heads tend to learn similar linear projections, resulting in redundancy in attention maps. The analysis shows that explicitly decomposing the computation of each head by feeding them with diverse features can mitigate this issue while improving {computation efficiency}.
In addition, the parameter allocation in different modules is often overlooked by existing lightweight models, as they mainly follow the configurations in standard transformer models \cite{deit, swin}. 
To improve parameter efficiency, we use structured pruning \cite{liu2018rethinking} to identify the most important network components, and summarize empirical guidance of {parameter reallocation} for model acceleration.

{Based upon the analysis and findings, we propose a new family of memory efficient transformer models named EfficientViT. 
Specifically, we design a new block with a sandwich layout to build up the model. The sandwich layout block applies a single memory-bound MHSA layer between FFN layers.} It reduces the time cost caused by memory-bound operations in MHSA, and applies more FFN layers to allow communication between different channels, which is more memory efficient. 
Then, we propose a new cascaded group attention (CGA) module to improve computation efficiency. The core idea is to enhance the diversity of the features fed into the attention heads. In contrast to prior self-attention using the same feature for all heads, CGA feeds each head with different input splits and cascades the output features across heads. 
This module not only reduces the computation redundancy in multi-head attention, but also elevates model capacity by increasing network depth. Last but not least, we redistribute parameters through expanding the channel width of critical network components such as value projections, while shrinking the ones with lower importance like hidden dimensions in FFNs. This reallocation finally promotes model parameter efficiency.

Experiments demonstrate that our models achieve clear improvements over existing efficient CNN and ViT models in terms of both speed and accuracy, as shown in Fig.~\ref{fig:model_gpu}. For instance, our {EfficientViT-M5} gets 77.1\% top-1 accuracy on ImageNet with throughput of 10,621 images/s on an Nvidia V100 GPU and 56.8 images/s on an Intel Xeon E5-2690 v4 CPU @ 2.60GHz, outperforming MobileNetV3-Large \cite{mobilenetv3} by 1.9\% in accuracy, 40.4\% in GPU inference speed, and 45.2\% in CPU speed. Moreover, EfficientViT-M2 gets 70.8\% accuracy, surpassing MobileViT-XXS \cite{mobilevit} by 1.8\%, while running 5.8$\times$/3.7$\times$ faster on the GPU/CPU, and 7.4$\times$ faster when converted to ONNX \cite{onnx} format. When deployed on the mobile chipset, \emph{i.e.}, Apple A13 Bionic chip in iPhone 11, EfficientViT-M2 model runs 2.3$\times$ faster than MobileViT-XXS \cite{mobilevit} using the CoreML\cite{CoreMLTools}.

In summary, the contributions of this work are two-fold:
\begin{itemize}
    \vspace{-7pt}
    \item We present a systematic analysis on the factors that affect the inference speed of  vision transformers, deriving a set of guidelines for efficient model design.
    \vspace{-6pt}
    \item We design a new family of vision transformer models, which strike a good trade-off between efficiency and accuracy. The models also demonstrate good transfer ability on a variety of downstream tasks.
    
\end{itemize}

\begin{figure}
\vspace{-12pt}
    \centering
    \includegraphics[width=0.433\textwidth]{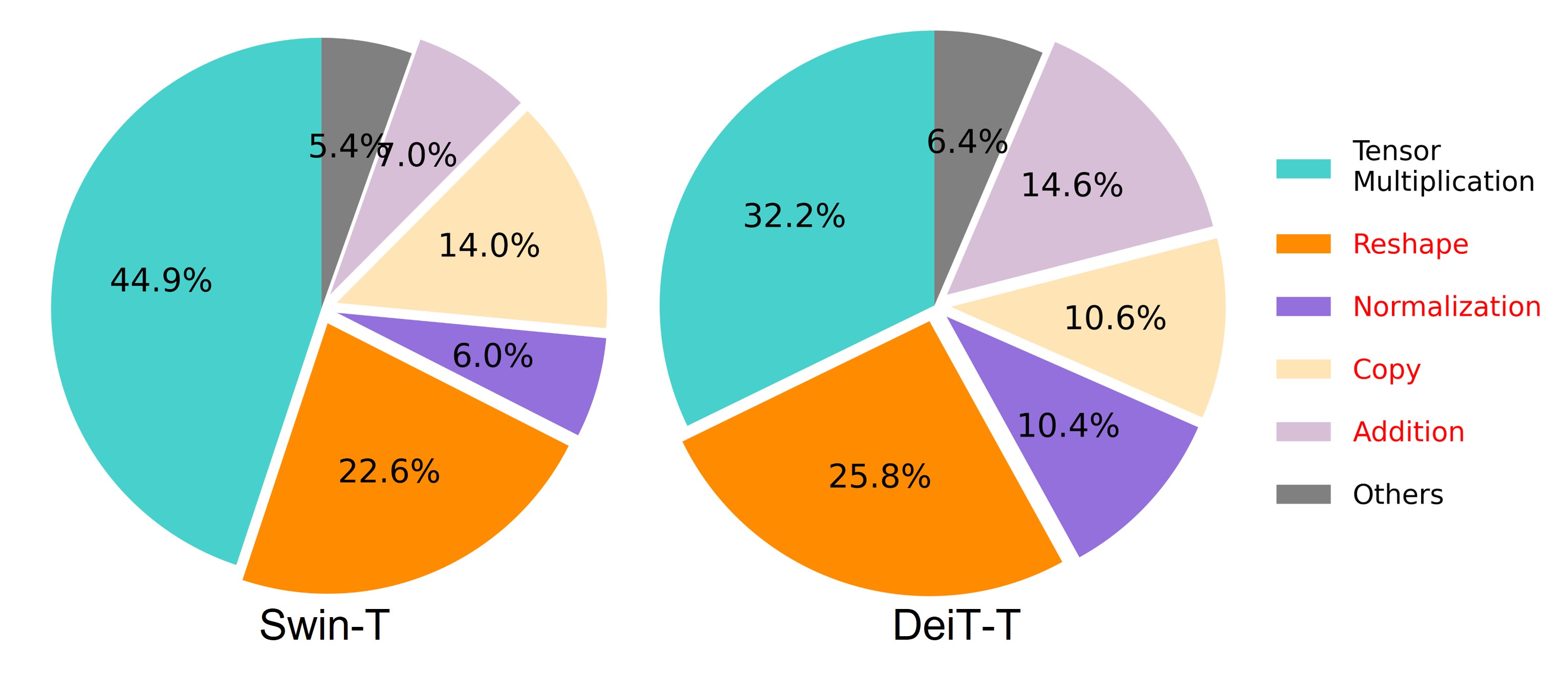}
    \vspace{-10pt}
    \caption{Runtime profiling on two standard vision transformers Swin-T and DeiT-T. \textcolor{red}{Red text} denotes memory-bound operations, \textit{i.e.}, the time taken by the operation is mainly determined by memory accesses, while time spent in computation is much smaller.}
    \label{fig:memory_compare_intro}
    \vspace{-15pt}
\end{figure}

\vspace{-12pt}
\section{Going Faster with Vision Transformers}
\vspace{-3pt}
In this section, we explore how to improve the efficiency of vision transformers from three perspectives: memory access, computation redundancy, and parameter usage. We seek to identify the underlying speed bottlenecks through empirical studies, and summarize useful design guidelines.

\vspace{-1pt}
\subsection{Memory Efficiency}
\label{sec:memeff}
\vspace{-3pt}

Memory access overhead is a critical factor affecting model speed \cite{dao2022flashattention, ivanov2021datamovement, tabani2021improving, hua2022transformerqualityinlineartime}. Many operators in transformer \cite{vaswani2017attention}, such as frequent reshaping, element-wise addition, and normalization are memory inefficient, requiring time-consuming access across different memory units, as shown in Fig. \ref{fig:memory_compare_intro}. Although there are some methods proposed to address this issue by simplifying the computation of standard softmax self-attention, \emph{e.g.}, sparse attention \cite{kitaev2020reformer, pvt, ren2021combiner, pan2022hilo} and low-rank approximation \cite{choromanski2020performer, wang2020linformer, mehta2022mobilevitv2}, they often come at the cost of accuracy degradation and limited acceleration.

\begin{figure}
\vspace{-8pt}
    \centering
    \includegraphics[width=0.421\textwidth]{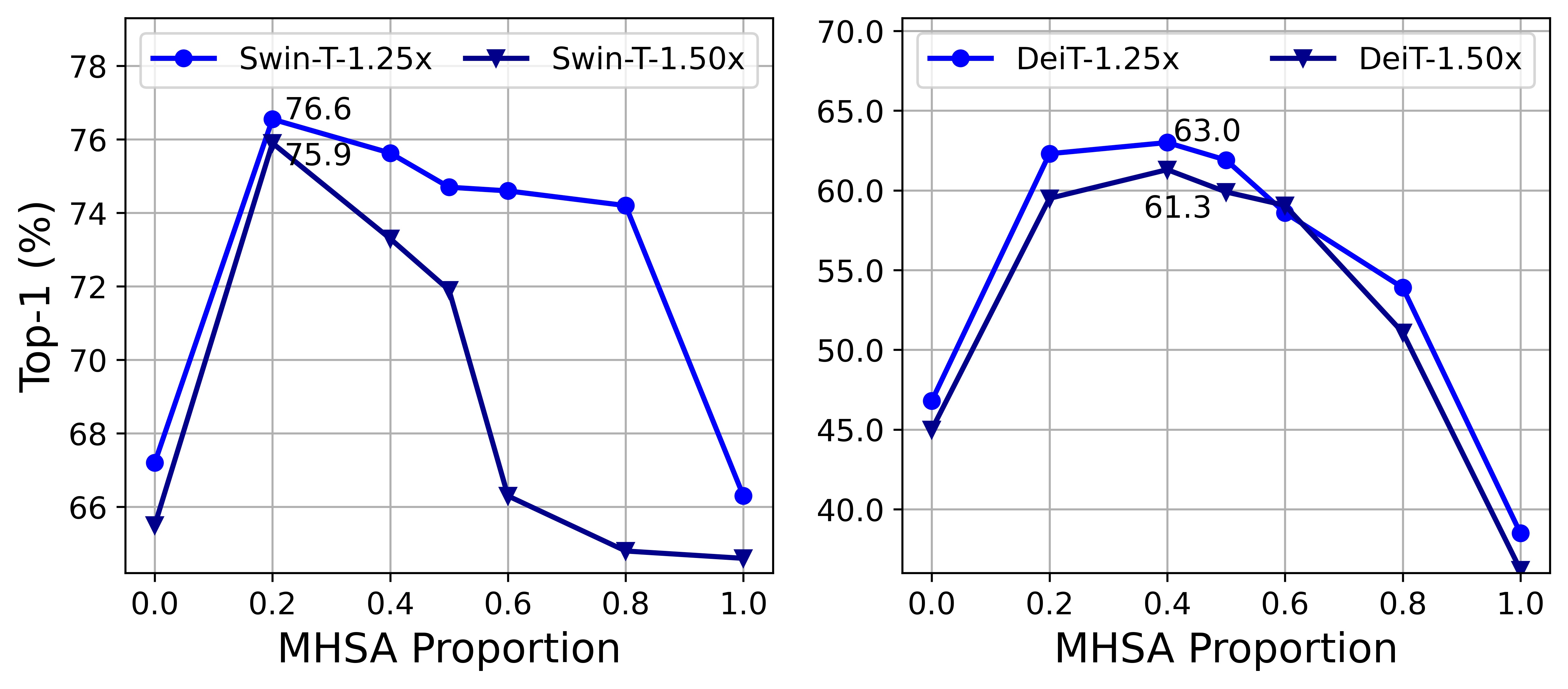}
    \vspace{-11pt}
    \caption[Caption for LOF]{The accuracy of downscaled baseline models with different MHSA layer proportions, where the dots on each line represent subnetworks with similar throughput.
    \textbf{Left:} Swin-T as the baseline. \textbf{Right:} DeiT-T as the baseline. 
    The 1.25$\times$/1.5$\times$ denote accelerating the
    baseline models by 1.25/1.5 times, respectively. 
    } 
    \label{fig:layer_redistribution}
    \vspace{-16pt}
\end{figure}

In this work, we turn to save memory access cost by reducing memory-inefficient layers. 
Recent studies reveal that memory-inefficient operations are mainly located in MHSA rather than FFN layers \cite{ivanov2021datamovement, kao2021optimized}. However, most existing ViTs \cite{ViT, deit, swin} use an equivalent number of these two layers, which may not achieve the optimal efficiency. We thereby explore the optimal allocation of MHSA and FFN layers in small models with fast inference. 
Specifically, we scale down Swin-T \cite{swin} and DeiT-T \cite{deit} to several small subnetworks with 1.25$\times$ and 1.5$\times$ higher inference throughput, and compare the performance of subnetworks with different proportions of MHSA layers.
As shown in Fig. \ref{fig:layer_redistribution}, subnetworks with 20\%-40\% MHSA layers tend to get better accuracy. 
Such ratios are much smaller than the typical ViTs that adopt 50\% MHSA layers. {Furthermore, we measure the time consumption on memory-bound operations to compare memory access efficiency,
including reshaping, element-wise addition, copying, and normalization.} 
{Memory-bound operations is reduced to 44.26\% of the total runtime in Swin-T-1.25$\times$ that has 20\% MHSA layers. }
The observation also generalizes to DeiT and smaller models with 1.5$\times$ speed-up. It is demonstrated that \textit{reducing MHSA layer utilization ratio appropriately can enhance memory efficiency while improving model performance.}

\vspace{-1pt}
\subsection{Computation Efficiency}
\label{sec:compeff}
\vspace{-2pt}

\begin{figure}
\vspace{-11pt}
    \centering
    \includegraphics[width=0.421\textwidth]{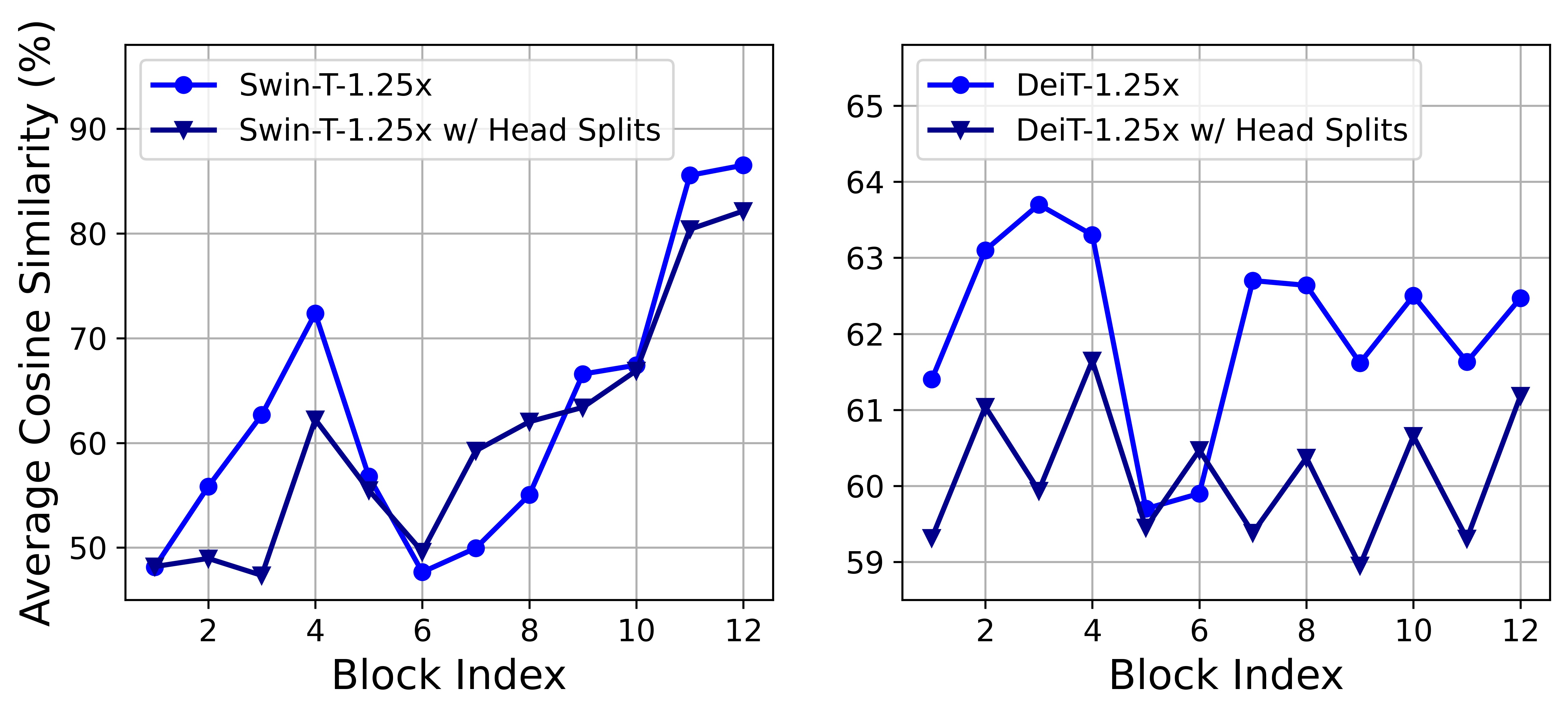}
    \vspace{-12pt}
    \caption{The average maximum cosine similarity of each head in different blocks. \textbf{Left:} downscaled Swin-T models. \textbf{Right:} downscaled DeiT-T models. Blue lines denote Swin-T-1.25$\times$/DeiT-T-1.25$\times$ model, while darkblue lines denote the variants that feed each head with only a split of the full feature.} 
    \label{fig:cosine_sim}
    \vspace{-8pt}
\end{figure}

MHSA embeds the input sequence into multiple subspaces (heads) and computes attention maps separately, which has been proven effective in improving performance \cite{vaswani2017attention, ViT, deit}. 
However, attention maps are computationally expensive, and studies have shown that a number of them are not of vital importance \cite{michel2019sixteen, voita2019analyzingmhsa_specialiedheadsdotheheavy}. 
To save computation cost, we explore how to reduce redundant attention in small ViT models. 
We train width downscaled Swin-T \cite{swin} and DeiT-T \cite{deit} models with 1.25$\times$ inference speed-up, and measure the maximum cosine similarity of each head and the remaining heads within each block. 
From Fig. \ref{fig:cosine_sim}, we observe there exists high similarities between attention heads, especially in the last blocks. The phenomenon suggests that many heads learn similar projections of the same full feature and incur computation redundancy.
To explicitly encourage the heads to learn different patterns, we apply an intuitive solution by feeding each head with only a split of the full feature, which is similar to the idea of group convolution in \cite{shufflenet, xception}.  
We train the variants of downscaled models
with the modified MHSA, and also compute the attention similarities in Fig. \ref{fig:cosine_sim}. It is shown that \textit{using different channel-wise splits of the feature in different heads, instead of using the same full feature for all heads as MHSA, could effectively mitigate attention computation redundancy.}

\begin{figure}
\vspace{-6pt}
    \centering
    \includegraphics[width=0.421\textwidth]{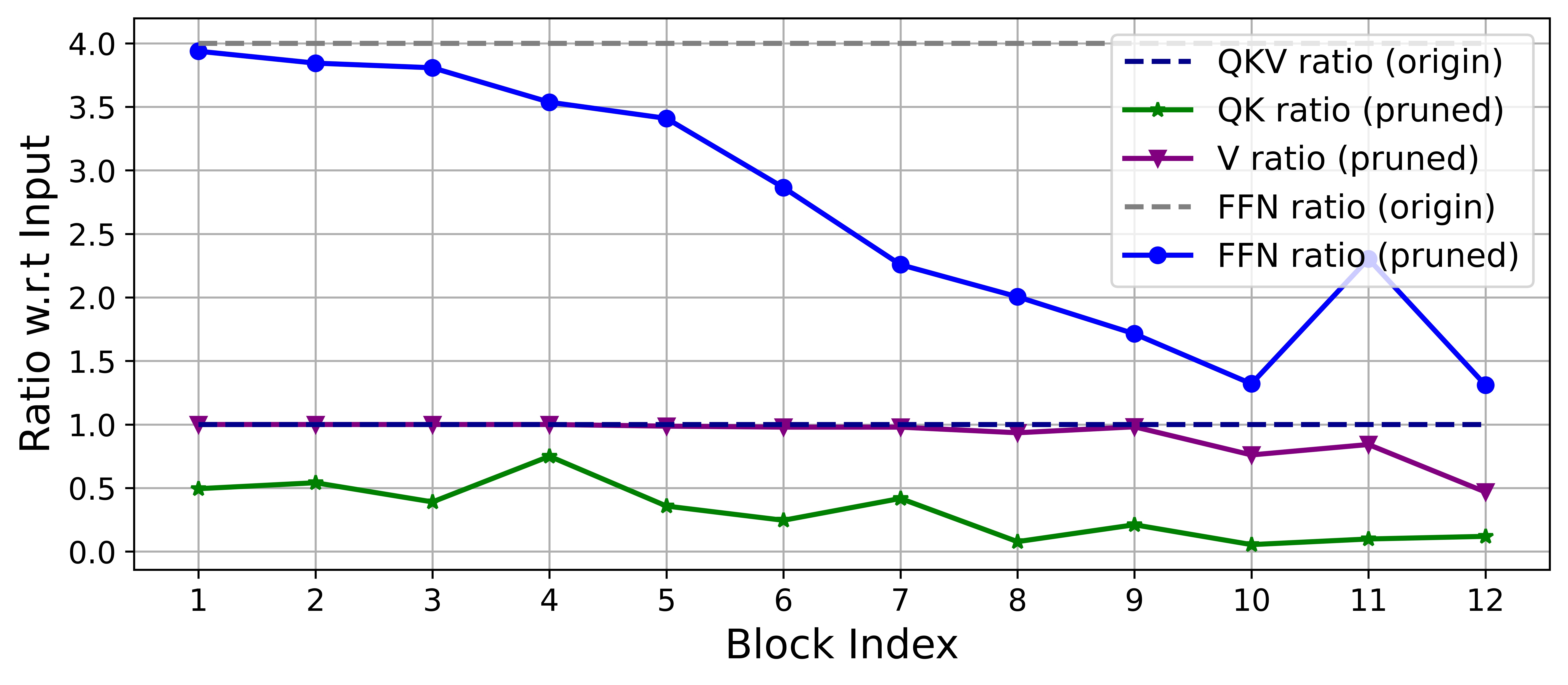}
    \vspace{-12pt}
    \caption{The ratio of the channels to the input embeddings before and after pruning Swin-T. 
    Baseline accuracy: 79.1\%; pruned accuracy: 76.5\%. Results for DeiT-T are given in the supplementary. 
    }
    \label{fig:pruned_model_ratio}
    \vspace{-15pt}
\end{figure}
\vspace{-1pt}
\subsection{Parameter Efficiency}
\label{sec:paraeff}
\vspace{-2pt}

Typical ViTs mainly inherit the design strategies from NLP transformer \cite{vaswani2017attention}, \emph{e.g.}, using an equivalent width for $Q$,$K$,$V$ projections, increasing heads over stages, and setting the expansion ratio to 4 in FFN. 
For lightweight models, the configurations of these components need to be carefully re-designed \cite{autoformer, li2022ds, autoscaling}. 
Inspired by \cite{liu2018rethinking, NViT}, we adopt Taylor structured pruning \cite{molchanov2019importance} to automatically find the important components in Swin-T and DeiT-T, and explore the underlying principles of parameter allocation. The pruning method removes unimportant channels under a certain resource constraint and keeps the most critical ones to best preserve the accuracy. It uses the multiplication of gradient and weight as channel importance, which approximates the loss fluctuation when removing channels \cite{lecun1989optimal}. 

The ratio between the remaining output channels to the input channels is plotted in Fig. \ref{fig:pruned_model_ratio}, and the original ratios in the unpruned model
are also given for reference. It is observed that: 1) The first two stages
preserve more dimensions, while the last stage
keeps much less; 2) The $Q$,$K$ and FFN dimensions are largely trimmed, whereas the dimension of $V$ is almost preserved and diminishes only at the last few blocks. These phenomena show that \textit{1    ) the typical channel configuration, that doubles the channel after each stage \cite{swin} or use equivalent channels for all blocks \cite{deit}, may produce substantial redundancy in last few blocks;
2) The redundancy in $Q$,$K$ is much larger than $V$ when they have the same dimensions. $V$ prefers a relative large channels, being close to the input embedding dimension.}

\begin{figure*}
\vspace{-14pt}
    \centering
    \includegraphics[width=0.82\textwidth]{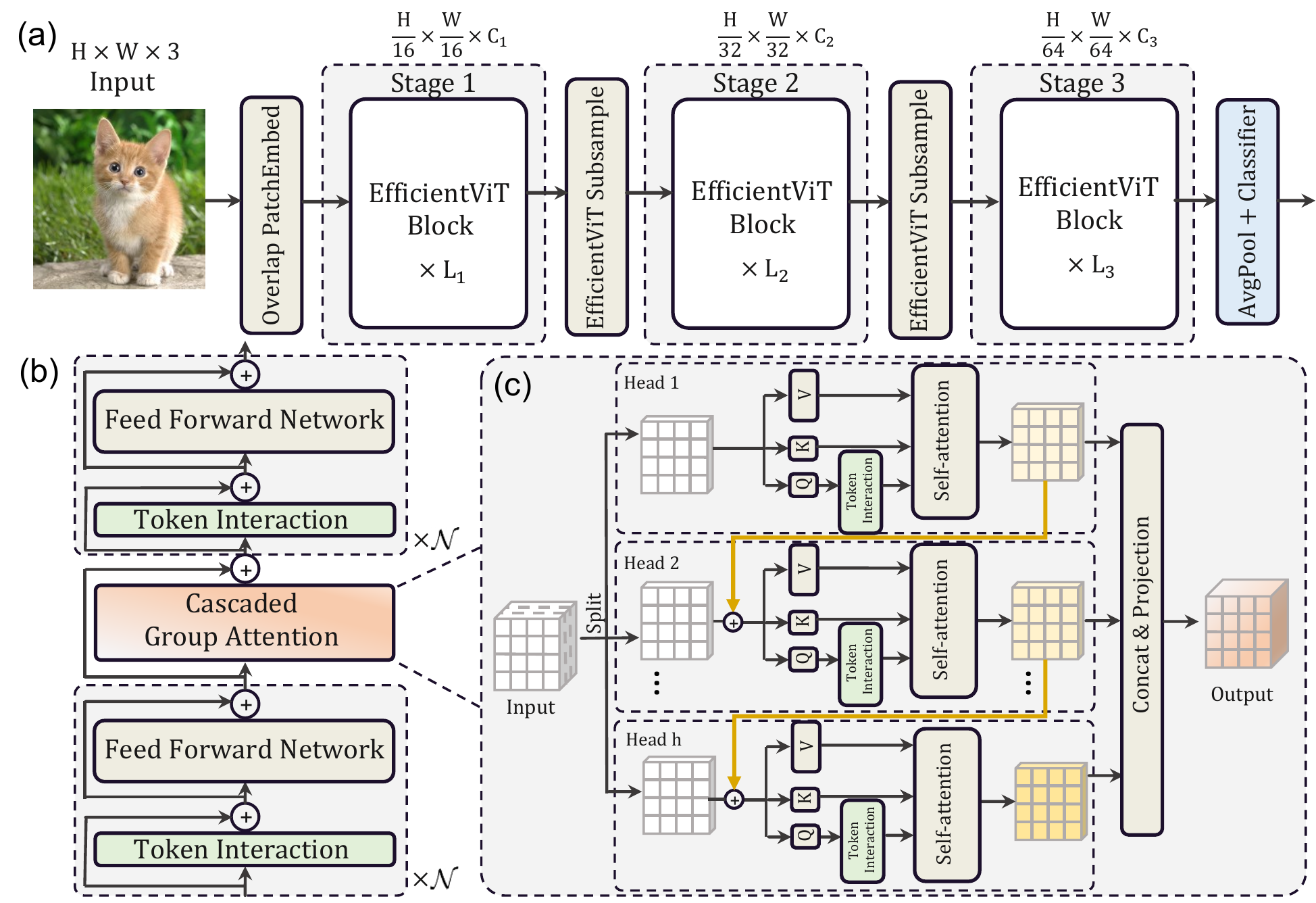}
    \vspace{-8pt}
    \caption{{Overview of EfficientViT. (a) Architecture of EfficientViT; (b) Sandwich Layout block; (c) Cascaded Group Attention.}} 
    \label{fig:arch}
    \vspace{-8pt}
\end{figure*}

\vspace{-6pt}
\section{Efficient Vision Transformer}
\vspace{-2pt}

Based upon the above analysis, in this section, we propose a new hierarchical model with fast inference named {EfficientViT}. The architecture overview is shown in Fig. \ref{fig:arch}.

\vspace{-1pt}
\subsection{EfficientViT Building Blocks}

We propose a new efficient building block for vision transformer, as shown in Fig. \ref{fig:arch} (b). It is composed of a memory-efficient sandwich layout, a cascaded group attention module, and a parameter reallocation strategy, which focus on improving model efficiency in terms of memory, computation, and parameter, respectively.

\textit{Sandwich Layout.} 
To build up a memory-efficient block, we propose a sandwich layout that employs less memory-bound self-attention layers and more memory-efficient FFN layers for channel communication. Specifically, it applies a single self-attention layer $\Phi^{\rm A}_i$ for spatial mixing, which is sandwiched between FFN layers $\Phi^{\rm F}_i$. The computation can be formulated as:
\vspace{-8pt}
\begin{equation}
\begin{aligned}
    X_{i+1} = \prod^{\mathcal{N}}\Phi_i^{\rm F} (\Phi_i^{\rm A}(\prod^{\mathcal{N}}\Phi_i^{\rm F}(X_i))),\\
\end{aligned}
\vspace{-4pt}
\end{equation}
where ${X}_i$ is the full input feature for the $i$-th block. The block transforms ${X}_i$ into ${X}_{i+1}$ with $\mathcal{N}$ FFNs before and after the single self-attention layer. 
This design reduces the memory time consumption caused by self-attention layers in the model, and applies more FFN layers to allow communication between different feature channels efficiently. We also apply an extra token interaction layer before each FFN using a depthwise convolution (DWConv) \cite{howard2017mobilenets}. It introduces {inductive bias of the local structural information} to enhance model capability \cite{dai2021coatnet}.

\textit{Cascaded Group Attention.} 
Attention head redundancy is a severe issue in MHSA, which causes computation inefficiency. Inspired by group convolutions in efficient CNNs \cite{alexnet, xception, shufflenet, inception}, we propose a new attention module named cascaded group attention (CGA) for vision transformers. It feeds each head with different splits of the full features, thus explicitly decomposing the attention computation across heads.
Formally, this attention can be formulated as:
\vspace{-5pt}
\begin{equation}
\begin{aligned}
\vspace{-9pt}
    {\widetilde{X}}_{ij} &= Attn(X_{ij}{W}^{\rm Q}_{ij}, X_{ij}{W}^{\rm K}_{ij}, X_{ij}{W}^{\rm V}_{ij}), \\
    {\widetilde{X}}_{i+1} &= Concat[{\widetilde{X}}_{ij}]_{j=1:h}{W}^{\rm P}_{i},
\end{aligned}
\label{eq2}
\vspace{-5pt}
\end{equation}
where the $j$-th head computes the self-attention over ${{X}}_{ij}$, which is the $j$-th split of the input feature ${{X}}_{i}$, \emph{i.e.}, ${{X}}_{i}$ $=$ $[{{X}}_{i1},{{X}}_{i2},\ldots, {{X}}_{ih} ]$ and $1$ $\leq$ $j$ $\leq$ $h$.  
$h$ is the total number of heads, ${W}^{\rm Q}_{ij}$, ${W}^{\rm K}_{ij}$, and ${W}^{\rm V}_{ij}$ are projection layers mapping the input feature split into different subspaces, and ${W}^{\rm P}_{i}$ is a linear layer that projects the concatenated output features back to the dimension consistent with the input. 

Although using feature splits instead of the full features for each head is more efficient and saves computation overhead, we continue to improve its capacity, by encouraging the $Q$, $K$, $V$ layers to learn projections on features with richer information. We compute the attention map of each head in a cascaded manner, as illustrated in Fig. \ref{fig:arch} (c), which adds the output of each head to the subsequent head to refine the feature representations progressively:
\vspace{-4pt}
\begin{equation}
    {X}^{'}_{ij} = X_{ij} + {\widetilde{X}}_{i(j-1)}, \quad 1 < j \le h,
\vspace{-7pt}
\end{equation}
where ${X}^{'}_{ij}$ is the addition of the $j$-th input split $X_{ij}$ and the $(j$$-$$1)$-th head output ${\widetilde{X}}_{i(j-1)}$ calculated by Eq. (\ref{eq2}). It replaces $X_{ij}$ to serve as the new input feature for the $j$-th head when calculating the self-attention. {Besides, another token interaction layer is applied after the $Q$ projection, which enables the self-attention to jointly capture local and global relations and further enhances the feature representation.}

\begin{table}[t]
  \centering
  \vspace{-8pt}
  \caption{Architecture details of EfficientViT model variants.}
  \vspace{-8pt}
\setlength\tabcolsep{1.3pt}
    \scalebox{0.82}{\begin{tabular}{l|ccc}
    \toprule
    Model & \{$C_1$, $C_2$, $C_3$\} & \{$L_1$, $L_2$, $L_3$\} & \{$H_1$, $H_2$, $H_3$\} \\
    \midrule
    EfficientViT-M0~~ & \{64, 128, 192\} & \{1, 2, 3\} & \{4, 4, 4\} \\
    EfficientViT-M1 & \{128, 144, 192\} & \{1, 2, 3\} & \{2, 3, 3\} \\
    EfficientViT-M2 & \{128, 192, 224\} & \{1, 2, 3\} & \{4, 3, 2\} \\
    EfficientViT-M3 & \{128, 240, 320\} & \{1, 2, 3\} & \{4, 3, 4\} \\
    EfficientViT-M4 & \{128, 256, 384\} & \{1, 2, 3\} & \{4, 4, 4\} \\
    EfficientViT-M5 & \{192, 288, 384\} & \{1, 3, 4\} & \{3, 3, 4\} \\
    \bottomrule
    \end{tabular}}%
    \vspace{-16pt}
  \label{tab:modelvariant}%
\end{table}%

\begin{table*}[htbp]
\vspace{-6pt}
\small
  \centering
  \caption{EfficientViT image classification performance on ImageNet-1K \cite{imagenet} with comparisons to state-of-the-art efficient CNN and ViT models trained without extra data. Throughput is tested on Nvidia V100 for GPU and Intel Xeon E5-2690 v4 @ 2.60 GHz processor for CPU and ONNX, where larger throughput means faster inference speed. $\uparrow$: finetune with higher resolution.}
  \vspace{-8pt}
    \begin{tabular}{l|cc|ccc|cccc}
    \toprule
    \multirow{2}[2]{*}{Model} & Top-1 & Top-5 & \multicolumn{3}{c|}{Throughput (images/s)}  & Flops & Params & \multirow{2}[2]{*}{Input} & \multirow{2}[2]{*}{Epochs} \\
\cmidrule{4-6}          & (\%) & (\%) & GPU   & CPU   & ONNX  & (M)   & (M)   &       &  \\
\midrule
    \rowcolor[rgb]{ .949,  .949,  .949} \textbf{EfficientViT-M0} & \textbf{63.2} & 85.4 & 27644 & 228.4  & 340.1   & 79 & 2.3    & 224   & 300 \\
    \midrule
    MobileNetV3-Small \cite{mobilenetv3} & 67.4 & - & 19738 & 156.5  & 231.7  & 57  & 2.5    & 224   & 600 \\
    \rowcolor[rgb]{ .949,  .949,  .949} \textbf{EfficientViT-M1} & \textbf{68.4} & 88.7 & 20093 & 126.9  & 215.9    & 167 & 3.0   & 224   & 300 \\
    Mobile-Former-52M \cite{mobileformer}& 68.7 & - & 3141  & 32.8  & 21.5   & 52 & 3.5     & 224   & 450 \\
    MobileViT-XXS \cite{mobilevit}& 69.0 & - & 4456  & 29.4  & 41.7  & 410  & 1.3    & 256   & 300 \\
    ShuffleNetV2 1.0$\times$ \cite{ma2018shufflenetv2}	&69.4& 88.9	&	13301	&106.7	&177.0		&146&2.3	&224	&300 \\
    MobileViTV2-0.5 \cite{mehta2022mobilevitv2}& 70.2 & - & 5142  & 34.4  & 44.9   & 466 & 1.4    & 256   & 300 \\
    \rowcolor[rgb]{ .949,  .949,  .949} \textbf{EfficientViT-M2} & \textbf{70.8} & 90.2 & 18218 & 121.2  & 158.7  & 201 & 4.2    & 224   & 300 \\
    \midrule
    MobileOne-S0 \cite{mobileone} & 71.4 & - & 11320 & 67.4 & 128.6 & 274 & 2.1 & 224 & 300 \\
    MobileNetV2 1.0$\times$ \cite{sandler2018mobilenetv2}& 72.0 & 91.0 & 6534  & 32.5  & 80.4  & 300  & 3.4    & 224   & 300 \\
    \rowcolor[rgb]{ .949,  .949,  .949} \textbf{EfficientViT-M3} & \textbf{73.4} & 91.4 & 16644 & 96.4  & 120.8  & 263 & 6.9    & 224   & 300 \\
    GhostNet 1.0$\times$ \cite{han2020ghostnet}& 73.9 & 91.4 & 7382  & 57.3  & 77.0    & 141 & 5.2   & 224   & 300 \\
    NASNet-A-Mobile \cite{nasnet} & 74.1 & - & 2623 & 19.8 & 25.5 & 564 & 5.3 & 224 & 300 \\ 
    \rowcolor[rgb]{ .949,  .949,  .949} \textbf{EfficientViT-M4} & \textbf{74.3} & 91.8 & 15914 & 88.5  & 108.6   & 299& 8.8    & 224   & 300 \\
    \midrule
    EdgeViT-XXS \cite{pan2022edgevits}& 74.4 & - & 3638  & 28.2  & 29.6    & 556 & 4.1   & 224   & 300 \\
    MobileViT-XS \cite{mobilevit}& 74.7 & - & 3344  & 11.1  & 20.5     & 986 & 2.3  & 256   & 300 \\
    ShuffleNetV2 2.0$\times$ \cite{ma2018shufflenetv2}	&74.9 & 92.4 & 6962&			37.9& 	52.3&	591&	7.4&	224&	300\\
    MobileNetV3-Large \cite{mobilenetv3}& 75.2 & - & 7560  & 39.1  & 70.5   & 217 & 5.4    & 224   & 600 \\
    MobileViTV2-0.75 \cite{mehta2022mobilevitv2}& 75.6 & - & 3350  & 16.0  & 22.7    & 1030 & 2.9  & 256   & 300 \\
    MobileOne-S1 \cite{mobileone} &	75.9 &-&	6663	&30.7	&51.1&	825	&4.8	&224 & 300 \\			
    GLiT-Tiny \cite{glit}& 76.4 & -& 3516&17.5&15.7&	1333& 7.3	&224&	300
    \\
    EfficientNet-B0 \cite{efficientnet}&  77.1 & 93.3 & 4532  & 30.2  & 29.5    & 390 & 5.3   & 224   & 350 \\
    \rowcolor[rgb]{ .949,  .949,  .949} \textbf{EfficientViT-M5} & \textbf{77.1} & 93.4 & 10621 & 56.8  & 62.5    & 522 & 12.4  & 224   & 300 \\
    \midrule
    \rowcolor[rgb]{ .949,  .949,  .949} \textbf{EfficientViT-M4$\uparrow$384} & \textbf{79.8} & 95.0 & 3986  & 15.8  & 22.6    & 1486 & 12.4 & 384   & 330 \\
    \rowcolor[rgb]{ .949,  .949,  .949} \textbf{EfficientViT-M5$\uparrow$512} & \textbf{80.8} & 95.5 & 2313  & 8.3   & 10.5    & 2670 & 12.4 & 512   & 360 \\
    \bottomrule
    \end{tabular}%
  \label{tab:imageclassification}%
  \vspace{-5pt}
\end{table*}%

Such a cascaded design enjoys two advantages. First, feeding each head with different feature splits could improve the diversity of attention maps, as validated in Sec. \ref{sec:compeff}. 
Similar to group convolutions \cite{shufflenet, xception}, the cascaded group attention could save the Flops and parameters by $h$$\times$, since the input and output channels in the $Q$$K$$V$ layers are reduced by $h$$\times$.
Second, cascading the attention heads allows for an increase of network depth, thus further elevating the model capacity without introducing any extra parameters. It only incurs minor latency overhead since the attention map computation in each head uses smaller $Q$$K$ channel dimensions.

\textit{Parameter Reallocation.}
To improve parameter efficiency, we reallocate the parameters in the network by expanding the channel width of critical modules while shrinking the unimportant ones. Specifically, based on the Taylor importance analysis in Sec. \ref{sec:paraeff}, we set small channel dimensions for $Q$ and $K$ projections in each head for all stages. For the $V$ projection, we allow it to have the same dimension as the input embedding. The expansion ratio in FFN is also reduced from 4 to 2 due to its parameter redundancy. With the proposed reallocation strategy, the important modules have larger number of channels to learn representations in a high dimensional space, which prevent the loss of feature information. Meanwhile, the redundant parameters in unimportant modules are removed to speed up inference and enhance the model efficiency.

\subsection{EfficientViT Network Architectures}
\vspace{-2pt}
The overall architecture of our EfficientViT is presented in Fig. \ref{fig:arch} (a).
Concretely, we introduce overlapping patch embedding \cite{early_conv, levit} to embed 16$\times$16 patches into tokens with $C_1$ dimension, which enhances the model capacity in low-level visual representation learning. 
The architecture contains three stages. Each stage stacks the proposed EfficientViT building blocks and the number of tokens is reduced by 4$\times$ at each subsampling layer (2$\times$ subsampling of the resolution). To achieve efficient subsampling, we propose an EfficientViT subsample block which also has the sandwich layout, except that the self-attention layer is replaced by an inverted residual block to reduce the information loss during subsampling 
\cite{sandler2018mobilenetv2, mobilenetv3}. %
It is worth noting that we adopt BatchNorm (BN) \cite{BN} instead of LayerNorm (LN) \cite{LN} throughout the model, as BN can be folded into the preceding convolution or linear layers, which is a runtime advantage over LN. We also use ReLU \cite{nair2010rectified} as the activation function, as the commonly used GELU \cite{gelu} or HardSwish \cite{mobilenetv3} are much slower, and sometimes not well-supported by certain inference deployment platforms \cite{onnx, CoreMLTools}.

We build our model family with six different width and depth scales, and set different number of heads for each stage. We use fewer blocks in early stages than late stages similar to MobileNetV3 \cite{mobilenetv3} and LeViT \cite{levit}, since that the processing on early stages with larger resolutions is more time consuming. We increase the width over stages with a small factor ($\le$ 2) to alleviate redundancy in later stages, as analyzed in Sec. \ref{sec:paraeff}. 
The architecture details of our model family are presented in Tab. \ref{tab:modelvariant}. $C_i$, $L_i$, and $H_i$ refer to the width, depth, and number of heads in the $i$-th stage. 

\vspace{-5pt}
\section{Experiments}
\subsection{Implementation Details}
\vspace{-2pt}
We conduct image classification experiments on ImageNet-1K \cite{imagenet}. The models are built with PyTorch 1.11.0 \cite{pytorch} and Timm 0.5.4 \cite{timm}, and trained from scratch for 300 epochs on 8 Nvidia V100 GPUs using AdamW \cite{adamw} optimizer and cosine learning rate scheduler. We set the total batchsize as 2,048. The input images are resized and randomly cropped into 224$\times$224. The initial learning rate is 1$\times$$10^{-3}$ with weight decay of 2.5$\times$$10^{-2}$. We use the same data augmentation as \cite{deit}, including Mixup \cite{mixup}, auto-augmentation \cite{autoaug}, and random erasing \cite{random_erase}. In addition, we provide throughput evaluation on different hardware. For GPU, we measure the throughput on an Nvidia V100, with the maximum power-of-two batchsize that fits in memory following \cite{deit, levit}. For CPU and ONNX, we measure the runtime on an Intel Xeon E5-2690 v4 @ 2.60 GHz processor, with batchsize 16 and run the model in a single thread following \cite{levit}. 
We also test the transferability of EfficientViT on downstream tasks. For the experiments on downstream image classification, we finetune the models for 300 epochs following \cite{MiniViT}, using AdamW \cite{adamw} with batchsize 256, learning rate 1$\times 10^{-3}$ and weight-decay 1$\times 10^{-8}$. We use RetinaNet \cite{retinanet} for object detection on COCO \cite{coco}, and train the models for 12 epochs (1$\times$ schedule) with the same settings as \cite{swin} on mmdetection \cite{chen2019mmdetection}. 
For instance segmentation, please refer to the supplementary.

\subsection{Results on ImageNet}
\vspace{-2pt}

We compare EfficientViT with prevailing efficient CNN and ViT models on ImageNet \cite{imagenet}, and report the results in Tab. \ref{tab:imageclassification} and Fig. \ref{fig:model_gpu}. The results show that, in most cases, our EfficientViT achieves the best accuracy and speed trade-off across different evaluation settings. 

\textit{Comparisons with efficient CNNs.} 
We first compare EfficientViT with 
vanilla CNN models, such as MobileNets \cite{mobilenetv3, sandler2018mobilenetv2} and EfficientNet \cite{efficientnet}. Specifically, compared to MobileNetV2 1.0$\times$ \cite{sandler2018mobilenetv2}, {EfficientViT-M3} obtains 1.4\% better top-1 accuracy, while running at 2.5$\times$ and 3.0$\times$ faster speed on V100 GPU and Intel CPU, respectively. Compared to the state-of-the-art MobileNetV3-Large \cite{mobilenetv3}, EfficientViT-M5 achieves 
1.9\% higher accuracy yet runs much faster, \emph{e.g.}, 40.5\% faster on the V100 GPU and 45.2\% faster on the Intel CPU but is 11.5\% slower as ONNX models. This may because reshaping is slower in ONNX implementation, which is inevitable in computing self-attention. Moreover, EfficientViT-M5 achieves 
comparable accuracy with the searched model EfficientNet-B0 \cite{efficientnet},
while runs 2.3$\times$/1.9$\times$ faster on the V100 GPU/Intel CPU, and 2.1$\times$ faster as ONNX models. Although our model uses more parameters, it reduces memory-inefficient operations that affect the inference speed and achieves higher throughput.

\vspace{-2pt}
\textit{Comparisons with efficient ViTs.} We also compare our models with recent efficient vision transformers \cite{mobileformer, mobilevit, mehta2022mobilevitv2, pan2022edgevits, glit} in Tab. \ref{tab:imageclassification}. 
In particular, when getting similar performance on ImageNet-1K \cite{imagenet}, our EfficientViT-M4 runs 4.4$\times$ and 3.0$\times$ faster than the recent EdgeViT-XXS \cite{pan2022edgevits} on the tested CPU and GPU devices. Even converted to ONNX runtime format, our model still gets 3.7$\times$ higher speed. Compared to the state-of-the-art MobileViTV2-0.5 \cite{mehta2022mobilevitv2}, our EfficientViT-M2 achieves slightly better performance with higher throughput, \emph{e.g.}, 3.4$\times$ and 3.5$\times$ higher throughput tested on the GPU and CPU devices, respectively. Furthermore, we compare with tiny variants of state-of-the-art large ViTs in Tab. \ref{tab:largecompare}. PoolFormer-12S \cite{yu2022metaformer} has comparable accuracy with EfficientViT-M5 yet runs 3.0$\times$ slower on the V100 GPU. Compared to Swin-T \cite{swin}, EfficientViT-M5 is 4.1\% inferior in accuracy yet is 12.3$\times$ faster on the Intel CPU, demonstrating the efficiency of the proposed design. In addition, we present the speed evaluation and comparison on mobile chipsets in the supplementary material.

\textit{Finetune with higher resolutions.} Recent works on ViTs have demonstrated that finetuning with higher resolutions can further improve the capacity of the models. We also finetune our largest model EfficientViT-M5 to higher resolutions. EfficientViT-M5$\uparrow$384 reaches 79.8\% top-1 accuracy with throughput of 3,986 images/s on the V100 GPU, and EfficientViT-M5$\uparrow$512 further improves the top-1 accuracy to 80.8\%, demonstrating the efficiency on processing images with larger resolutions and the good model capacity. 

\begin{table}[t]
\vspace{-6pt}
  \centering
  \caption{Comparison with the tiny variants of state-of-the-art large-scale ViTs on ImageNet-1K \cite{imagenet}.}
  \vspace{-8pt}
    \scalebox{0.71}{\begin{tabular}{l|c|ccc|cc}
    \toprule
    \multirow{2}[4]{*}{Model} & Top-1 & \multicolumn{3}{c|}{Throughput (imgs/s)} & Flops & Params \\
\cmidrule{3-5}          & (\%)  & GPU   & CPU   & ONNX  & (G)   & (M) \\
    \midrule
    PVTV2-B0 \cite{pvt_v2}& 70.5  & 3507  & 12.7  & 18.5  & 0.6   & 1.4 \\
    T2T-ViT-7 \cite{T2TViT} & 71.7 & 1156 & 22.5 & 16.1  & 1.1& 4.3\\
    DeiT-T \cite{deit}& 72.2  & 4631  & 26.0    & 25.1  & 1.3   & 5.9 \\
    PoolFormer-12S \cite{yu2022metaformer}& 77.2  & 3534  & 10.4  & 14.6  & 1.9   & 12.0 \\
    EffFormer-L1 \cite{li2022efficientformer} & 79.2  & 4465  & 12.9  & 21.2  & 1.3   & 12.3 \\
    Swin-T \cite{swin}& \textbf{81.2}  & 1393  & 4.6   & 6.4   & 4.5   & 29.0 \\
    \midrule
    \rowcolor[rgb]{ .949,  .949,  .949}\textbf{EfficientViT-M5} & 77.1  & \textbf{10621} & \textbf{56.8}  & \textbf{62.5}  & 0.5   & 12.4 \\
    \bottomrule
    \end{tabular}}%
    \vspace{-10pt}
  \label{tab:largecompare}%
\end{table}%

\begin{table}[t]
  \centering
  \caption{Results of EfficientViT and other efficient models on downstream image classification datasets.}
  \vspace{-8pt}
    \scalebox{0.75}{\begin{tabular}{l|c|c|ccccc}
    \toprule
    Model & \rotatebox[origin=l]{90}{\footnotesize{Throughput}} & \rotatebox[origin=l]{90}{\footnotesize{ImageNet}} & \rotatebox[origin=l]{90}{\footnotesize{CIFAR10}} & \rotatebox[origin=l]{90}{\footnotesize{CIFAR100}} & \rotatebox[origin=l]{90}{\footnotesize{Cars}} & \rotatebox[origin=l]{90}{\footnotesize{Flowers}} & \rotatebox[origin=l]{90}{\footnotesize{Pets}} \\
    \midrule
    MobileNetV1 \cite{howard2017mobilenets}& 8543  & 70.6  & 96.1  & 82.3  & \textbf{91.4}  & 96.7  & 89.9 \\
    MobileNetV2 \cite{sandler2018mobilenetv2}& 6534  & 72.9  & 95.7  & 80.8  & 91.0    & 96.6  & 90.5 \\
    MobileNetV3 \cite{mobilenetv3}& 7560  & 75.2  & 97.6  & 85.5  & 91.2    & 97.0  & 90.1
    \\
    NASNet-A-M \cite{nasnet}& 2623  & 74.1  & 96.8  & 83.9  & 88.5  & 96.8  & 89.4 \\
    ViT-S/16 \cite{ViT}& 2135  & \textbf{81.4}  & 97.6  & 85.7  & -     & 86.4  & 90.4 \\
    \rowcolor[rgb]{ .949,  .949,  .949}\textbf{EfficientViT-M5} & \textbf{10621} & {77.1}  & \textbf{98.0}    & \textbf{86.4}  & 89.7  & \textbf{97.1}  & \textbf{92.0} \\
    \bottomrule
    \end{tabular}}%
  \label{tab:transfer}%
  \vspace{-15pt}
\end{table}%

\subsection{Transfer Learning Results}
\vspace{-2pt}

To further evaluate the transfer ability, we apply EfficientViT on various downstream tasks. 

\textit{Downstream Image Classification.} We transfer EfficientViT to downstream image classification datasets to test its generalization ability: 1) CIFAR-10 and CIFAR-100 \cite{cifar}; 2) fine-grained classification: Flowers \cite{flowers}, Stanford Cars \cite{cars}, and Oxford-IIIT Pets \cite{pets}. We report the results in Tab. \ref{tab:transfer}. Compared to existing efficient models \cite{howard2017mobilenets, sandler2018mobilenetv2, mobilenetv3, nasnet, ViT}, our EfficientViT-M5 achieves comparable or slightly better accuracy across all datasets with much higher throughput. An exception lies in Cars, where our model is slightly inferior in accuracy. This may because the subtle differences between classes lie more in local details thus is more feasible to be captured with convolution.

\textit{Object Detection.} We compare EfficientViT-M4 with efficient models \cite{sandler2018mobilenetv2, mobilenetv3, guo2020spos, tan2019mnasnet, chu2021fairnas, tan2019mixconv} on the COCO \cite{coco} object detection task, and present the results in Tab. \ref{tab:objectdetection}. Specifically, EfficientViT-M4 surpasses MobileNetV2 \cite{sandler2018mobilenetv2} by 4.4\% AP with comparable Flops. Compared to the searched method SPOS \cite{guo2020spos}, our EfficientViT-M4 uses 18.1\% fewer Flops while achieving 2.0\% higher AP, demonstrating its capacity and generalization ability in different vision tasks.

\subsection{Ablation Study}
In this section, we ablate important design elements in the proposed EfficientViT on ImageNet-1K \cite{imagenet}. 
All models are trained for 100 epochs to magnify the differences and reduce training time \cite{levit}. Tab. \ref{tab:ablation} reports the results.

\begin{table}[t]
\vspace{-6pt}
  \centering
  \caption{EfficientViT object detection performance on COCO \texttt{val2017} \cite{coco} with comparisons to other efficient models. 
  } 
  \vspace{-8pt}
\setlength\tabcolsep{1.8pt}
    \scalebox{0.77}{\begin{tabular}{l|cccccc|cc}
    \toprule
    \multirow{2}[4]{*}{Model} & \multicolumn{6}{c|}{RetinaNet 1$\times$} & Flops & \multicolumn{1}{c}{Params} \\
\cmidrule{2-7}          & ${\rm AP}$    & ${\rm AP}_{50}$  & ${\rm AP}_{75}$  & ${\rm AP}_{s}$   & ${\rm AP}_{m}$   & ${\rm AP}_{l}$  & (M)   & \multicolumn{1}{c}{(M)} \\
    \midrule
    MobileNetV2 \cite{sandler2018mobilenetv2} ~~~~& 28.3 & 46.7 & 29.3 & 14.8 & 30.7 & 38.1 & 300 & 3.4\\
    MobileNetV3 \cite{mobilenetv3}& 29.9 &49.3 &30.8 &14.9 &33.3 &41.1 & 217 & 5.4 \\
    SPOS \cite{guo2020spos} &  30.7 & 49.8 & 32.2 & 15.4 &33.9 &41.6 & 365 & 4.3\\
    MNASNet-A2 \cite{tan2019mnasnet}& 30.5 &50.2 &32.0 &16.6 &34.1 &41.1 & 340 & 4.8 \\
    FairNAS-C \cite{chu2021fairnas} &  31.2 &50.8 &32.7 &16.3 &34.4 &42.3 & 325 & 5.6 \\
    MixNet-M \cite{tan2019mixconv}  & 31.3 &51.7 &32.4 &17.0 &35.0 &41.9 & 360 & 5.0 \\
    \rowcolor[rgb]{ .949,  .949,  .949} \textbf{EfficientViT-M4} & \textbf{32.7} & \textbf{52.2} & \textbf{34.1} & \textbf{17.6} & \textbf{35.3} & \textbf{46.0} & 299 & 8.8 \\
    \bottomrule
    \end{tabular}}%
  \label{tab:objectdetection}%
  \vspace{-5pt}
\end{table}%

\begin{table}[t]
\vspace{-6pt}
  \centering
  \caption{Ablation for EfficientViT-M4 on ImageNet-1K \cite{imagenet} dataset. Top-1 accuracy, GPU and ONNX throughput are reported.}
  \vspace{-6pt}
    \scalebox{0.76}{\begin{tabular}{cl|c|cc}
    \toprule
    \multirow{2}[4]{*}{\#} & \multirow{2}[4]{*}{Ablation} & \multicolumn{1}{c|}{\multirow{2}[4]{*}{Top-1 (\%)}} & \multicolumn{2}{c}{Throughput (imgs/s)} \\
\cmidrule{4-5}          &    &   & ~GPU   & ONNX \\
    \midrule
    \rowcolor[rgb]{ .949,  .949,  .949} 1 &EfficientViT-M4 & 71.3  & ~15914 & 108.6  \\
    \midrule
    2&Sandwich → Swin \cite{swin} & 68.3  & ~15804 & 114.5 \\
    3&$\mathcal{N}$ = 1 → 2 & 70.2 & ~14977 & 112.3 \\
    4&$\mathcal{N}$ = 1 → 3 & 65.7 & ~15856 & 139.7 \\
    \midrule
    5&CGA → MHSA \cite{ViT}& 70.2  & ~16243 & 102.2 \\
    6&Cascade → None & 69.8  & ~16411 & 111.0 \\
    \midrule
    7&QKV allocation → None & 69.9 & ~15132 & 103.1\\
    8&FFN ratio 2 → 4 & 69.8 & ~15310 & 112.4 \\
    \midrule
    9&DWConv → None & 69.9 & ~16325 & 110.4 \\
    10&BN → LN \cite{LN}& 70.4                & ~15463& 103.6 \\
    11&ReLU → HSwish \cite{mobilenetv3} & 72.2  & ~15887 & 87.5  \\
    \bottomrule
    \end{tabular}}%
  \label{tab:ablation}%
  \vspace{-18pt}
\end{table}%

\textit{Impact of the sandwich layout block.} We first present an ablation study to verify the efficiency of the proposed sandwich layout design, by replacing the sandwich layout block with the original Swin block \cite{swin}. The depth is adjusted to \{2, 2, 3\} to guarantee similar throughput with EfficientViT-M4 for a fair comparison. 
The top-1 accuracy degrades by 3.0\% at a similar speed, verifying that applying more FFNs instead of memory-bound MHSA is more effective for small models.
Furthermore, to analyze the impact of the number of FFNs $\mathcal{N}$ before and after self-attention
, we change the number from 1 to 2 and 3. The number of blocks is reduced accordingly to maintain similar throughput. As presented in Tab. \ref{tab:ablation} (\#3 and \#4), 
further increasing the number of FFNs is not effective due to the lack of long-range spatial relation
and $\mathcal{N}$=1 achieves the best efficiency.

\textit{Impact of the cascaded group attention.} We have proposed CGA to improve the computation efficiency of MHSA. As shown in Tab. \ref{tab:ablation} (\#5 and \#6), replacing CGA with MHSA decreases the accuracy by 1.1\% and ONNX speed by 5.9\%, suggesting that addressing head redundancy improves the model efficiency. For the model without the cascade operation, its performance is comparable with MHSA but worse than CGA, demonstrating the efficacy of enhancing the feature representations of each head.

\begin{figure}
\vspace{-12pt}
    \centering
    \includegraphics[width=0.47\textwidth]{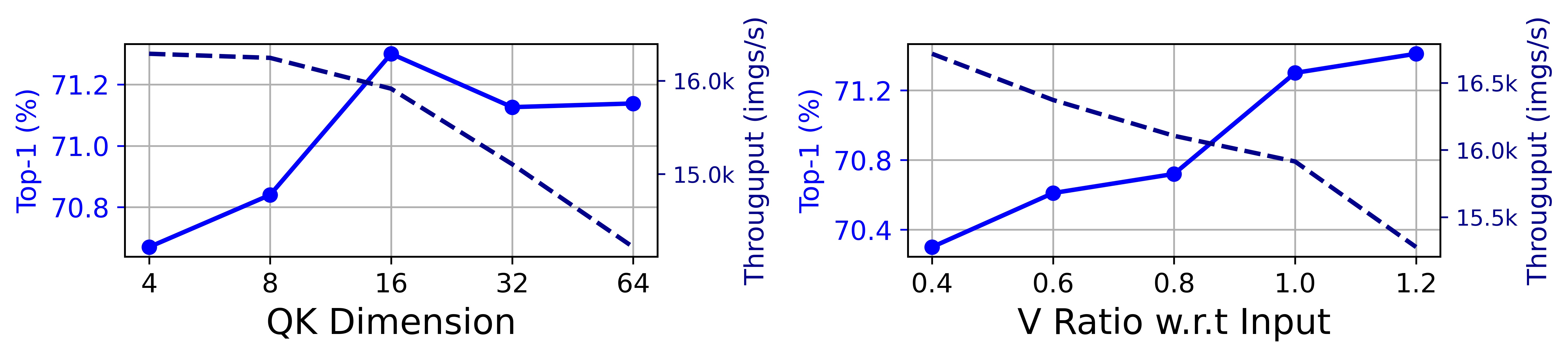}
    \vspace{-12pt}
    \caption{Ablation on the $Q$$K$ dimension of each head and the ratio of $V$ dimension to the input embedding.}
    \vspace{-16pt}
    \label{fig:qkv}
\end{figure}

\textit{Impact of the parameter reallocation.} 
Our EfficientViT-M4 yields 1.4\%/1.5\% higher top-1 accuracy, 4.9\%/3.8\% higher GPU throughput than the models without $Q$$K$$V$ channel dimension reallocation or FFN ratio reduction, respectively, indicating the effectiveness of parameter reallocation (\#1 \textit{vs.} \#7, \#8). Moreover, we study the choices of $Q$$K$ dimension in each head and the ratio of $V$ dimension to the input embedding in Fig. \ref{fig:qkv}. {
It is shown that the performance is improved gradually as $Q$$K$ dimension increases from 4 to 16,} while further increasing it gives inferior performance. 
Besides, the performance improves from 70.3\% to 71.3\% when increasing the ratio 
between $V$ dimension and input embedding from 0.4 to 1.0. When further enlarging the ratio to 1.2, it only gets 0.1\% improvements. Therefore, setting the channels of $V$ close to the input embedding achieves the best parameter efficiency, which meets our analysis in Sec. \ref{sec:paraeff} and design strategy.

\textit{Impact of other components.} 
We ablate the impact of using DWConv for token interaction, the normalization layer, and the activation function, as presented in Tab. \ref{tab:ablation} (\#9, \#10, and \#11). With DWConv, the accuracy improves by 1.4\% with a minor latency overhead, demonstrating the effectiveness of introducing local structural information. Replacing BN with LN decreases accuracy by
0.9\% and GPU speed by 2.9\%.
Using HardSwish instead of ReLU
improves accuracy by 0.9\% but leads to a large drop of 20.0\% ONNX speed. The activation functions are element-wise operations that occupy a considerable amount of processing time on {GPU/CPU} \cite{ma2018shufflenetv2, dao2022flashattention, venkat2019swirl},
thus utilizing ReLU instead of more complicated activation functions is of better efficiency.

\begin{table}[t]
  \centering
\vspace{-7pt}
  \caption{Performance comparison on ImageNet-1K \cite{imagenet} and ImageNet-ReaL \cite{imagenet_real}. 
  Results with $\dag$ are trained with 1,000 epochs and knowledge distillation following LeViT \cite{levit}.}
  \vspace{-11pt}
  \setlength\tabcolsep{1.2pt}
    \scalebox{0.76}{\begin{tabular}{l|ccc|ccc|cc}
    \toprule
    \multirow{2}[4]{*}{Model} & \multicolumn{3}{c|}{ImageNet (\%)} & \multicolumn{3}{c|}{Throughput (imgs/s)} & Flops & Params \\
\cmidrule{2-7}          & Top-1 & Top-1$^\dag$ & ReaL$^\dag$  & GPU   & CPU   & ONNX  & (M)   & (M) \\
    \midrule
    LeViT-128S \cite{levit} & 73.6  & 76.6  & 82.6  & 14457 & 82.3  & 80.9  & 305   & \textbf{7.8} \\
    \rowcolor[rgb]{ .949,  .949,  .949} \textbf{EfficientViT-M4} & \textbf{74.3} & \textbf{77.1} & \textbf{83.6} & \textbf{15914} & \textbf{88.5} & \textbf{108.6} & \textbf{299} & 8.8 \\
    \bottomrule
    \end{tabular}}%
  \label{tab:improved_recipe}%
  \vspace{-14pt}
\end{table}%
\textit{Results of 1,000 training epochs and distillation.} {Tab. \ref{tab:improved_recipe} shows the results with 1,000 training epochs and knowledge distillation using RegNetY-16GF \cite{radosavovic2020designing} as the teacher model 
following \cite{levit} on ImageNet-1K \cite{imagenet} and ImageNet-ReaL \cite{imagenet_real}. 
Compared to LeViT-128S \cite{levit}, EfficientViT-M4 surpasses it by 0.5\% on ImageNet-1K and 1.0\% on ImageNet-ReaL, respectively. For the inference speed, our model has 34.2\% higher throughput on ONNX and also shows superiority on other settings. The results demonstrate that the strong capability and generalization ability of EfficientViT can be further explored with longer training schedules.}

\vspace{-4pt}
\section{Related Work}
\vspace{-2pt}

\textit{Efficient CNNs.} With the demand of deploying CNNs on resource-constrained scenarios, efficient CNNs have been intensively studied in literature \cite{sandler2018mobilenetv2, mobilenetv3, ma2018shufflenetv2, efficientnet, shufflenet, han2020ghostnet, han2022ghostnets}. Xception \cite{xception} proposes an architecture built with depthwise separable convolutions. MobileNetV2 \cite{sandler2018mobilenetv2} builds an inverted residual structure which expands the input to a higher dimension. MobileNetV3 \cite{mobilenetv3} and EfficientNet \cite{efficientnet} 
resort to neural architecture search techniques to design compact models. To boost the actual speed on hardware, ShuffleNetV2 \cite{ma2018shufflenetv2} introduces channel split and shuffle operations to improve the information communication among channel groups. 
However, the spatial locality of convolutional kernels hampers CNN models from capturing long-range dependencies, thus limiting their model capacity. 

\textit{Efficient ViTs.} ViT and its variants \cite{ViT, deit, swin, pvt_v2} have achieved success on various vision tasks. Despite the superior performance,
most of them are inferior to typical CNNs in inference speed. Some efficient transformers have been proposed recently and they fall into two camps: 1) efficient self-attention; and 2) efficient architecture design. Efficient self-attention methods reduce the cost of softmax attention via sparse attention \cite{kitaev2020reformer, pvt, ren2021combiner, pan2022hilo} or low-rank approximation \cite{choromanski2020performer, wang2020linformer, mehta2022mobilevitv2}. However, they suffer from performance degradation with negligible or moderate inference acceleration over softmax attention \cite{vaswani2017attention}. Another line of work combines ViTs with lightweight CNNs to build efficient architectures \cite{lvt, mobilevit, mobileformer, maaz2022edgenext, luo2022towards}. 
LVT \cite{lvt} proposes enhanced attention mechanisms with dilated convolution to improve the model performance and efficiency. 
Mobile-Former \cite{mobileformer} designs a parallel CNN-transformer block to encode both local features and global interaction.
However, most of them target at minimizing 
Flops and parameters \cite{dehghani2021efficiencymisnomer}, which could have low correlations with actual inference latency \cite{mobileone} and still inferior to efficient CNNs in speed. Different from them, we explore models with fast inference by directly optimizing their throughput on different hardware and deployment settings, and design a family of hierarchical models with a good trade-off between speed and accuracy. 

\vspace{-6pt}
\section{Conclusion}
\vspace{-4pt}

In this paper, we have presented a systematic analysis on the factors that affect the inference speed of vision transformers, and proposed a new family of fast vision transformers with memory-efficient operations and cascaded group attention, named EfficientViT. Extensive experiments have demonstrated the efficacy and high speed of EfficientViT, and also show its superiority on various downstream benchmarks. 

\textit{Limitations}. One limitation of EfficientViT is that, despite its high inference speed, the model size is slightly larger compared to state-of-the-art efficient CNN \cite{mobilenetv3} due to the extra FFNs in the introduced sandwich layout. Besides, our models are designed manually based on the derived guidelines on building efficient vision transformers. In future work, we are interested in reducing the model size and incorporating automatic search techniques to further enhance the model capacity and efficiency.

{\normalsize \noindent\textbf{Acknowledgement.} Prof. Yuan was partially supported by Hong Kong Research Grants Council (RGC) General Research Fund 11211221, and Innovation and Technology Commission-Innovation and Technology Fund ITS/100/20.}

{\small
\bibliographystyle{ieee_fullname}
\bibliography{egbib}
}
\end{document}